\documentclass[10pt,journal,compsoc]{IEEEtran}

\usepackage{booktabs} %

\usepackage{multirow}
\usepackage{enumitem}

\usepackage{graphicx}

\usepackage{amsmath}
\usepackage{amssymb}
\usepackage{mathtools}
\usepackage{amsthm}

\usepackage{amsmath,amsfonts,bm}
\usepackage{xcolor}
\usepackage{fontawesome}

\def\eqref#1{(\ref{#1})}

\def\1{\bm{1}}

\def\vzero{{\bm{0}}}

\def\vmu{{\bm{\mu}}}

\def\vb{{\bm{b}}}

\def\vk{{\bm{k}}}

\def\vo{{\bm{o}}}

\def\vq{{\bm{q}}}

\def\vu{{\bm{u}}}
\def\vv{{\bm{v}}}

\def\vx{{\bm{x}}}
\def\vy{{\bm{y}}}
\def\vz{{\bm{z}}}

\def\mP{{\bm{P}}}

\def\mW{{\bm{W}}}
\def\mX{{\bm{X}}}

\def\mSigma{{\bm{\Sigma}}}

\DeclareMathAlphabet{\mathsfit}{\encodingdefault}{\sfdefault}{m}{sl}
\SetMathAlphabet{\mathsfit}{bold}{\encodingdefault}{\sfdefault}{bx}{n}

\def\gL{{\mathcal{L}}}

\def\gN{{\mathcal{N}}}

\def\gT{{\mathcal{T}}}

\def\gV{{\mathcal{V}}}

\def\sL{{\mathbb{L}}}

\def\sX{{\mathbb{X}}}

\newcommand{\R}{\mathbb{R}}

\DeclareMathOperator*{\argmin}{arg\,min}

\newcommand{\abs}[1]{\left\lvert #1 \right\rvert}
\newcommand{\ip}[2]{\left\langle#1,#2\right\rangle}

\newcommand{\T}{\textup{T}}

\newcommand{\rev}[1]{\textcolor{red}{#1}}

\theoremstyle{plain}
\newtheorem{theorem}{Theorem}[section]
\newtheorem{proposition}[theorem]{Proposition}

\theoremstyle{definition}

\theoremstyle{remark}

\renewcommand{\paragraph}{\vspace*{0.25em}\noindent\textbf}

\usepackage[numbers]{natbib}

\ifCLASSINFOpdf
\else
\fi

\begin{document}
\title{Hyperbolic Convolution via Kernel Point Aggregation}

\author{Eric~Qu,~Dongmian~Zou~\IEEEmembership{Member, IEEE}%
\IEEEcompsocitemizethanks{\IEEEcompsocthanksitem E. Qu was with the Department
of Electrical Engineering and Computer Science, University of California at Berkeley, CA 94720, USA.\protect\\
E-mail: ericqu@berkeley.edu
\IEEEcompsocthanksitem D. Zou was with CMCS, DSRC, Division of Natural and Applied Sciences, Duke Kunshan University, Kunshan, Jiangsu 215316, China.\protect\\
E-mail: dongmian.zou@duke.edu}%
\thanks{The research results of this article are sponsored by the Kunshan Municipal Government research funding.}
\thanks{(Corresponding author: Dongmian Zou.)}
}

\markboth{Journal of \LaTeX\ Class Files,~Vol.~14, No.~8, August~2015}%
{Qu \MakeLowercase{\textit{et al.}}: Hyperbolic Convolution via Kernel Point Aggregation}

\IEEEtitleabstractindextext{%
\begin{abstract}
Learning representations according to the underlying geometry is of vital importance for non-Euclidean data. Studies have revealed that the hyperbolic space can effectively embed hierarchical or tree-like data. In particular, the few past years have witnessed a rapid development of hyperbolic neural networks. However, it is challenging to learn good hyperbolic representations since common Euclidean neural operations, such as convolution, do not extend to the hyperbolic space. Most hyperbolic neural networks do not embrace the convolution operation and ignore local patterns. Others either only use non-hyperbolic convolution, or miss essential properties such as equivariance to permutation. We propose HKConv, a novel trainable hyperbolic convolution which first correlates trainable local hyperbolic features with fixed kernel points placed in the hyperbolic space, then aggregates the output features within a local neighborhood. HKConv not only expressively learns local features according to the hyperbolic geometry, but also enjoys equivariance to permutation of hyperbolic points and invariance to parallel transport of a local neighborhood. We show that neural networks with HKConv layers advance state-of-the-art in various tasks. 
\end{abstract}

}

\maketitle

\IEEEdisplaynontitleabstractindextext

\IEEEpeerreviewmaketitle

\section{Introduction}\label{sec:intro}

\IEEEPARstart{R}{ecent} advances of geometric deep learning have achieved remarkable successes for data in non-Euclidean domains.
In applications, many datasets show a hierarchical or tree-like structure. 
For these data, the hyperbolic space has proved effective as it has large capacity and allows embedding with small distortion \citep{sonthalia2020tree}.
Many hyperbolic neural network methods have arisen and found wide applications such as word embedding \citep{nagano2019wrapped}, knowledge graph completion \citep{chami2020low}, and image segmentation \citep{atigh2022hyperbolic}.
In those neural networks, Euclidean neural operations are generalized to one of the hyperbolic coordinate models, such as the Poincar\'{e} ball model or the Lorentz model. 

One core component in hyperbolic neural networks is the linear layers, known as fully-connected layers, in the hyperbolic space \citep{ganea2018hyperbolic, chami2019hyperbolic, shimizu2021hyperbolic, chen2021fully}. 
However, hyperbolic linear layers are unaware of the structure within the input data, especially local patterns.
In the Euclidean domain, convolution is often used to extract local information from the input. The local connectivity has made convolutional layers an important building block that fostered the early success of deep learning \citep{lecun2015deep}.
In convolutional neural networks (CNN), convolutional filters extract local patterns from the input by measuring and aggregating correlations between the filter and different regions of the input. 
Unfortunately, hyperbolic neural networks seldom contain convolutional layers. Some hyperbolic networks \citep{khrulkov2020hyperbolic, ahmad2022fisheyehdk, atigh2022hyperbolic} carry out convolutions in regular CNN layers, which are not regarded as hyperbolic convolutions. The only existing work that explicitly defines hyperbolic convolution is HNN++ \citep{shimizu2021hyperbolic}, where convolution is done by first concatenating features in the reception field and then applying a hyperbolic fully-connected layer. Although the convolution in HNN++ effectively incorporates neighborhood information, it could only apply to ordered and structured data (e.g., sequences or images). This is because it is not equivariant to permutation of the order of input features. Simply switching the order of concatenation will completely change the output. Moreover, it only applies to the case where each feature has the same number of neighbors and does not work for heterogeneous relations between data points. This makes it impossible to extend to graph input.

\IEEEpubidadjcol

On the other hand, many works \citep{liu2019hyperbolic, chami2019hyperbolic, bachmann2020constant, dai2021hyperbolic, zhang2021lorentzian} have generalized graph convolutional network (GCN) to the hyperbolic space. The main idea is to use message passing to extract local patterns defined by a graph. However, such graph convolution depends only on the topology of the underlying graph and neglects the rich geometric patterns in the hyperbolic space itself. 

One difficulty with designing hyperbolic convolution lies in the fact that principles in designing CNN convolutions do not extend to the hyperbolic space.
On one hand, the hyperbolic space is usually used as the embedding space of features, not a domain where the features are defined as functions. On the other hand, the hyperbolic space is not a regular grid and cannot facilitate alignment of convolutional kernels and input features. 
To address the former issue, the convolutional filter has to be defined also in the same hyperbolic space; to address the latter issue, the convolutional filter should not be required to swipe over the space.
In view of these requirements, we draw inspiration from point cloud analysis \citep{atzmon2018point, thomas2019kpconv, lin2020convolution} and use fixed point kernels in the hyperbolic space to design a hyperbolic convolutional layer, called the hyperbolic kernel convolution (HKConv). 
HKConv draws learnable transformations from relative positions of input features, whose correlation with the point kernels are then processed to produce the output features. 
We outline our contributions as follows.
\begin{itemize}[leftmargin=12pt, parsep=0pt]
    \item We formulate HKConv, a novel hyperbolic convolution method based on point kernels. To the best of our knowledge, this is the first hyperbolic neural layer that effectively extracts local hyperbolic patterns.
    \item HKConv is equivariant to permutation of ordering of the input hyperbolic features. Moreover, it is invariant to parallel translation of a local neighborhood.
    \item We demonstrate the effectiveness of HKConv by using it in hyperbolic neural networks, which achieve state-of-the-art performance in graph classification, node classification and machine translation. 
\end{itemize}

\section{Background}

\subsection{Hyperbolic Geometry}

Hyperbolic geometry is a non-Euclidean geometry with a constant negative curvature \citep{cannon1997hyperbolic, anderson2006hyperbolic}. In order to define operations in the hyperbolic space, it is convenient to work with a Cartesian-like model (coordinate system) such as the Lorentz model and the Poincaré ball model (both are isometric). We choose to work with the Lorentz model since it is known to be numerically stable and provide sufficient space for optimization \citep{nickel2018learning, chami2019hyperbolic}.

Given a constant negative curvature $\kappa < 0$, the Lorentz model $\sL^n = \mathbb{L}^n_\kappa=(\mathcal{L},\mathfrak{g})$, known as the hyperboloid, is an $n$-dimensional manifold $\mathcal{L}$ embedded in $\R^{n+1}$. 
Every point in $\mathbb{L}^n$ is represented by 
$\boldsymbol{x}=\begin{bmatrix}x_t\\ \boldsymbol{x}_s\end{bmatrix}$, where $x_t > 0$ is the ``time component'' and $\vx_s\in \mathbb{R}^n$ is the ``spatial component''. They satisfy $\ip{\vx}{\vx}_{\gL} = 1/\kappa$, where $\langle \cdot,\cdot\rangle_\mathcal{L}$ is the Lorentz inner product induced by the metric tensor $\mathfrak{g}=\operatorname{diag}([-1,\boldsymbol{1}_n^\top])$ ($\boldsymbol{1}_n \in \R^n$ is the $n$-dimensional vector whose entries are all $1$'s): $\langle \boldsymbol x,\boldsymbol y\rangle_\mathcal{L}\coloneqq\boldsymbol x^\top \mathfrak{g}\boldsymbol y=-x_t y_t +\boldsymbol x_s^\top \boldsymbol y_s, ~\vx, \vy \in \sL^n$.

\paragraph{Geodesics} Shortest paths in the hyperbolic space are called geodesics. The distance between two points in hyperbolic space is the length of the geodesic between them. With the Lorentz model,  the distance between two points $\vx$ and $\vy$ is given by, e.g., \citet{chami2019hyperbolic}:
\begin{equation}\label{eq:def_lorentz_dist}
    d_\mathcal{L}(\boldsymbol x,\boldsymbol y)={(-\kappa)}^{-1/2}\cosh^{-1}(\kappa\langle\boldsymbol x,\boldsymbol y\rangle_\mathcal{L}).
\end{equation}

\paragraph{Tangent Space} The tangent space at $\boldsymbol x \in \sL^n$, $\mathcal{T}_{\boldsymbol{x}}\mathbb{L}^n$, is the first order approximation of the manifold around $\vx$, and can be represented as an $n$-dimensional subspace of $\mathbb{R}^{n+1}$. 
Specifically, $\mathcal{T}_{\boldsymbol{x}}\mathbb{L}^n\coloneqq \{\boldsymbol{y}\in \mathbb{R}^{n+1}: \langle \boldsymbol y,\boldsymbol x\rangle_\mathcal{L}=0\}$.

\paragraph{Exponential and Logarithmic Maps} For $\boldsymbol x \in \mathbb{L}^n$, the exponential map $\operatorname{exp}_{\boldsymbol{x}}:\mathcal{T}_{\boldsymbol{x}}\mathbb{L}^n\to \mathbb{L}^n$ projects a tangent vector into the hyperbolic space. Let $\gamma$ be the geodesic satisfying $\gamma(0)=\boldsymbol x$ and $\gamma'(0)=\boldsymbol v$. Then $\operatorname{exp}_{\boldsymbol{x}}(\boldsymbol v)\coloneqq \gamma(1)$. 
With the Lorentz model, $$\operatorname{exp}_{\boldsymbol{x}}(\boldsymbol v)=\cosh(\phi)\boldsymbol x + \phi^{-1} \sinh(\phi){\boldsymbol v},\quad \phi=\sqrt{-\kappa}\|\boldsymbol v\|_\mathcal{L},$$
where $\|\boldsymbol \cdot \|_\mathcal{L} := \sqrt{\langle \cdot, \cdot \rangle_\mathcal{L}}$.
The logarithmic map $\operatorname{log}_{\boldsymbol{x}}: \mathbb{L}^n\to \mathcal{T}_{\boldsymbol{x}}\mathbb{L}^n$ projects hyperbolic vectors into the tangent space. It is the inverse map of $\exp_\vx$ in the sense that $\operatorname{log}_{\boldsymbol{x}}(\operatorname{exp}_{\boldsymbol{x}}(\boldsymbol v))=\boldsymbol v$ for any $\vv \in \sL^n$ and $\exp_\vx ( \log_\vx (\vu) ) = \vu$ for any $\vu \in \gT_\vx \sL^n$. Using Lorentz coordinates, 
\begin{equation*}
\operatorname{log}_{\boldsymbol{x}}(\boldsymbol u)=\frac{\cosh^{-1}(\psi)}{\sqrt{-\kappa}}\frac{\boldsymbol u-\psi \boldsymbol x}{\|\vu-\psi\vx\|_\mathcal{L}},\psi=\kappa \langle\boldsymbol x,\boldsymbol u\rangle_\mathcal{L}.
\end{equation*}

\paragraph{Parallel Translation} For two points $\boldsymbol{x},\boldsymbol y\in \mathbb{L}^n$, the parallel transport, also called the parallel translation, from $\boldsymbol x$ to $\boldsymbol y$, $\operatorname{PT}_{\boldsymbol x\to \boldsymbol y}$, is a map that ``transports'' or ``translates'' a tangent vector from $\mathcal{T}_{\boldsymbol{x}}\mathbb{L}^n$ to $\mathcal{T}_{\boldsymbol{y}}\mathbb{L}^n$ along the geodesic from $\boldsymbol x$ to $\boldsymbol y$. The parallel transport in $\sL^n$ is
$$\operatorname{PT}_{\boldsymbol x\to \boldsymbol y}(\boldsymbol v)=\frac{\langle \boldsymbol y, \boldsymbol v\rangle_\mathcal{L}}{-1/\kappa-\langle \boldsymbol x, \boldsymbol y\rangle_\mathcal{L}}(\boldsymbol x+ \boldsymbol y).$$

\subsection{Hyperbolic Neural Operations}\label{app:prelim}

\paragraph{Hyperbolic Linear Layers} Cascading many exponential and logarithmic maps may lead to numerical instability \citep{chami2019hyperbolic, chen2021fully}. Therefore, when designing linear layers in complex hyperbolic neural networks, it is beneficial to take a ``fully hyperbolic'' approach, originally proposed by \citet{chen2021fully}. A fully hyperbolic linear layer applies a linear transformation to the spatial component of the input, and fills the time component so that the whole vector lies in the hyperboloid. Specifically, it maps $\vx \in \sL^m$ to $\begin{bmatrix} {\sqrt{\|\boldsymbol{Wx}\|^2-1/\kappa}} \\ \boldsymbol W \vx\end{bmatrix}$, where $\mW \in \R^{n \times (m+1)}$.
With additional activation, bias and normalization, a general expressive linear layer transforms an input $\vx \in \sL^m$ to
$$\boldsymbol{y}=\operatorname{HLinear}(\boldsymbol{x})=\begin{bmatrix}\sqrt{\|h(\vx)\|^2-1/\kappa}\\h(\vx)\end{bmatrix} \in \sL^n.$$
Here,
\begin{equation*}
    h(\vx) = \frac{\lambda \sigma\left(\boldsymbol{v}^{\top} \boldsymbol{x}+b^{\prime}\right)}{\|\boldsymbol{W} \tau(\boldsymbol{x})+\boldsymbol{b}\|}(\boldsymbol{W} \tau(\boldsymbol{x})+\boldsymbol{b}),
\end{equation*}
where $\boldsymbol{v} \in \mathbb{R}^{n+1}$ and $\boldsymbol{W} \in \mathbb{R}^{n \times(m+1)}$ are trainable weights, $\boldsymbol{b} \in \R^n$ and $b^{\prime} \in \R$ are trainable biases, $\sigma$ is the sigmoid function, $\tau$ is the activation function, and the trainable parameter $\lambda>0$ scales the range. We may also write $h(\vx; \mW, \vb)$ to emphasize its dependence on $\mW$ and $\vb$.

\paragraph{Hyperbolic Centroid} The notion of a centroid is extended to $\sL^n$ by \citet{law2019lorentzian}, defined to be the point $\vmu^*$ that minimizes a weighted sum of squared Lorentzian distance: $$\vmu^* = \argmin_{\boldsymbol\mu \in \mathbb{L}^n}\sum_{i=1}^N\nu_i d_{\mathcal{L}}^2(\boldsymbol{x}_i,\boldsymbol\mu), $$ where $\{\vx_i\}_{i=1}^N$ is the set of points to aggregate, $\boldsymbol \nu$ is the weight vector whose entries satisfy $\nu_i\geq 0$, $\sum_i\nu_i>0$, $i = 1,\cdots, N$. A closed form of the centroid is given by
\begin{equation*}\label{eq:h_centroid}
 \operatorname{HCent}(\{\vx_i\}_{i=1}^N,\boldsymbol{\nu})= \boldsymbol\mu^* = 
 \frac{\sum_{i=1}^N\nu_i\boldsymbol{x}_i}{\sqrt{-\kappa}\left|\|\sum_{i=1}^N\nu_i\boldsymbol{x}_i\|_{\mathcal{L}}\right|}.
\end{equation*}

\rev{}

\paragraph{Hyperbolic Distance Layer} The hyperbolic distance layer \citep{liu2019hyperbolic} is a numerically stable way of output Euclidean features. It maps points from $\mathbb{L}^n$ to $\mathbb{R}^m$. Given an input $\boldsymbol{x}\in \mathbb{L}^n$, it first initializes $m$ trainable centroids $\{\boldsymbol{c}_i\}_{i=1}^m \subset \mathbb{L}^n$, then produces a vector of distances
\begin{equation*}\boldsymbol{y} = \operatorname{HCDist}_{n,m}(\boldsymbol{x})=[d_{\mathcal{L}}(\boldsymbol{x}, \boldsymbol{c}_1)  \cdots  d_{\mathcal{L}}(\boldsymbol{x}, \boldsymbol{c}_m)]^\top,\end{equation*}

\paragraph{Hyperbolic Embedding} A simple approach to embedding a vector $\vx \in \R^n$ in $\sL^n$ is as follows \citep{nagano2019wrapped}. First, a zero padding is added to the first coordinate of $\vx$, as if $\vx$ is the spatial component of $\hat{\vx} = \begin{bmatrix}0 \\ \vx \end{bmatrix} \in \gT_\vo \sL^n$, where $\gT_\vo \sL^n$ is the tangent space of $\sL^n$ at the hyperbolic origin $\vo$. Then, the embedding is produced by taking the exponential map of $\hat{\vx}$ at $\vo$.

\paragraph{Wrapped Normal Distribution} The wrapped normal distribution \citep{nagano2019wrapped} is a hyperbolic distribution that resembles the normal distribution in Euclidean space. Given $\boldsymbol \mu \in \mathbb{L}^n$ and $\mSigma \in \R^{n \times n}$, to sample $\vz\in\mathbb{L}^n$ from the wrapped normal distribution $\mathcal G(\boldsymbol \mu, \mSigma)$, one first samples a vector from the Euclidean normal distribution $\mathcal N(\boldsymbol 0, \mSigma)$ and applies an hyperbolic embedding to obtain $\vx \in \sL^n$. Then, one applies parallel transport and produces $\vz = \exp_\vmu \left( \operatorname{PT}_{\vo \to \vmu} \left( \log_\vo (\vx) \right) \right) \in \sL^n$. We used the wrapped normal distribution as an alternative way of generating the kernel points. However, our ablation study showed its deteriorate performance.

\section{Proposed Model}

\subsection{Hyperbolic Point Kernels}\label{sec:kp}

To construct an HKConv layer, the first step is to fix $K$ kernel points $\{\tilde{\vx}_k\}_{k=1}^K$ in $\sL^m$, where $m$ is the dimension of the input features. 
The kernel points can be regarded as located in the neighborhood of the hyperbolic origin $\vo = [1, \vzero_m^\T]^\T \in \sL^m$, where $\vzero_m$ is the $m$-dimensional zero vector.

The kernel points are pre-determined since it will be numerically unstable to train them in the hyperbolic space together with other parameters (see ablations in the experiments). This does not restrict the expressivity of HKConv, since learnable transformations are still applied to the input features.

The locations of $\{\tilde{\vx}_k\}_{k=1}^K$ are determined according to the principle of \citet{thomas2019kpconv}: the kernel points should be as far as possible from each other, but also not too distant from the origin. Specifically, the kernels are determined by solving the optimization problem where the following loss function is minimized:
\begin{equation}\label{eq:loss_ker}
    \mathfrak{L}(\{\tilde{\vx}_k\}_{k=1}^K) = \sum_{k=1}^K \sum_{l\not= k}{1\over d_{\gL}(\tilde{\vx}_l, \tilde{\vx}_k)} + \sum_{k=1}^K d_{\gL}(\vo, \tilde{\vx}_k).
\end{equation}

In our context, the first term in \eqref{eq:loss_ker} guarantees expressivity as it associates distinct features with each kernel point. If the kernel points are too close to each other, the output from the correlation of the input and the kernel points will be too similar. The second term ensures the solution does not diverge to infinity. Another reason for having the second term is to prevent vanishing gradients. When the kernel points are too distant, gradients of correlation may vanish easily because of the $\cosh^{-1}$ function used in $d_\gL$ (expressed in \eqref{eq:def_lorentz_dist}). We further illustrate a case of gradient vanishment in Appendix \ref{app:hkp}.

\begin{figure}[t]
  \centering
  \includegraphics[width = \columnwidth]{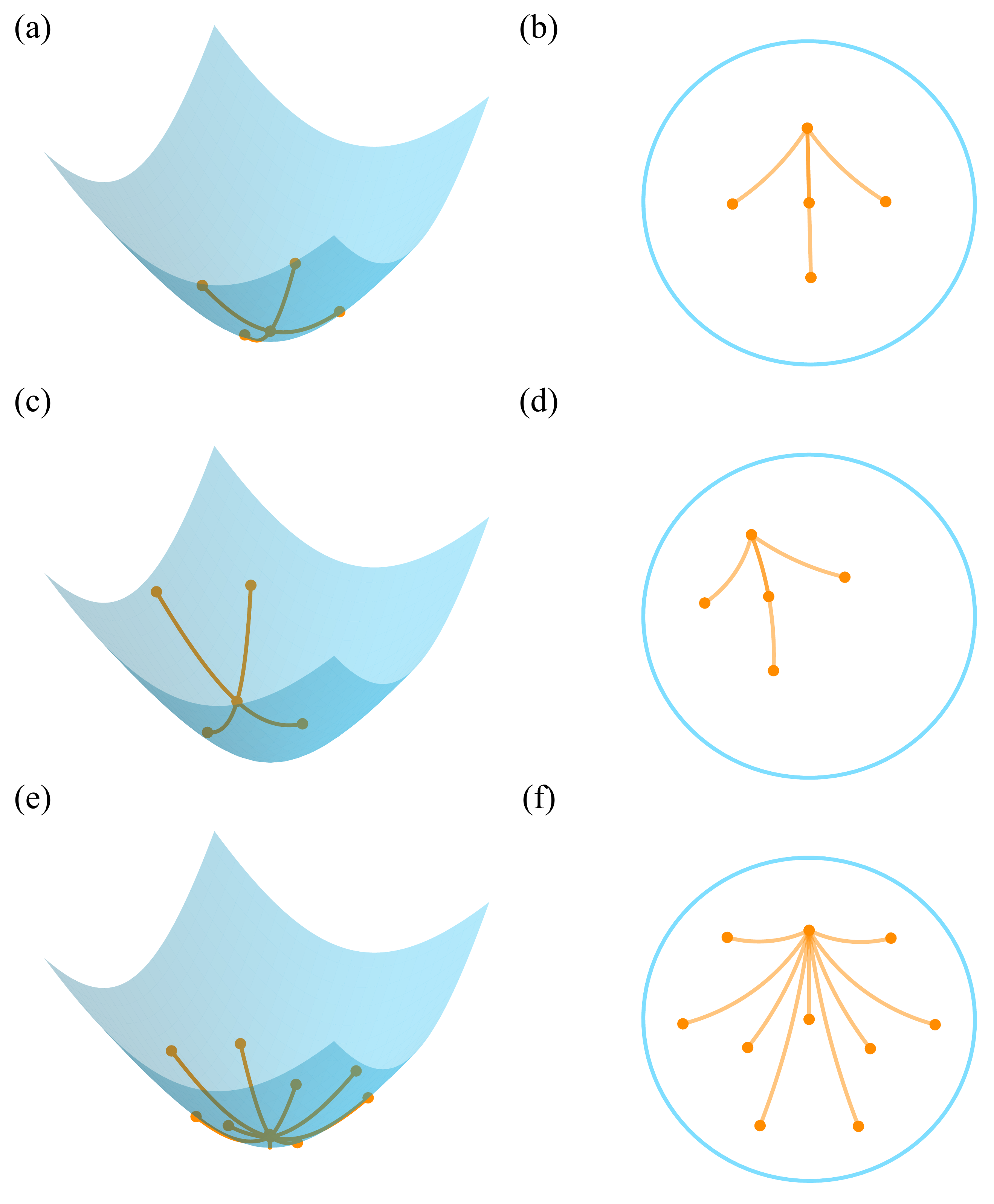}
  \caption{Illustration of kernel points in $\sL^2$. (a) and (b) illustrate $K=5$ points in the Lorentz and Poincar\'{e} models, respectively; (c) and (d) apply a parallel transport to all the points in (a) and (b); (e) and (f) show $K=10$ points in the Lorentz and Poincar\'{e} models, respectively. In each subfigure, we show the geodesics from one starting point to all the others (to show different patterns, we take two different choices of starting points in the Lorentz and Poincar\'{e} models, respectively).}
\label{fig:kp}
\end{figure}

In general, it is difficult to determine an analytical solution to the minimization of \eqref{eq:loss_ker}. We use the Riemannian gradient descent method \citep{becigneul2018riemannian} with a learning rate of $0.0001$ until it converges. Figure \ref{fig:kp} shows examples of $5$ and $10$ kernel points in $\sL^2$. We observe that the kernel points are around the origin; meanwhile, no pair of points are too close to each other. This agrees with our intuition in the formulation of \eqref{eq:loss_ker}.

\subsection{Hyperbolic Kernel Convolution 
}

\begin{figure}[t]
    \centering
    \includegraphics[width =\columnwidth]{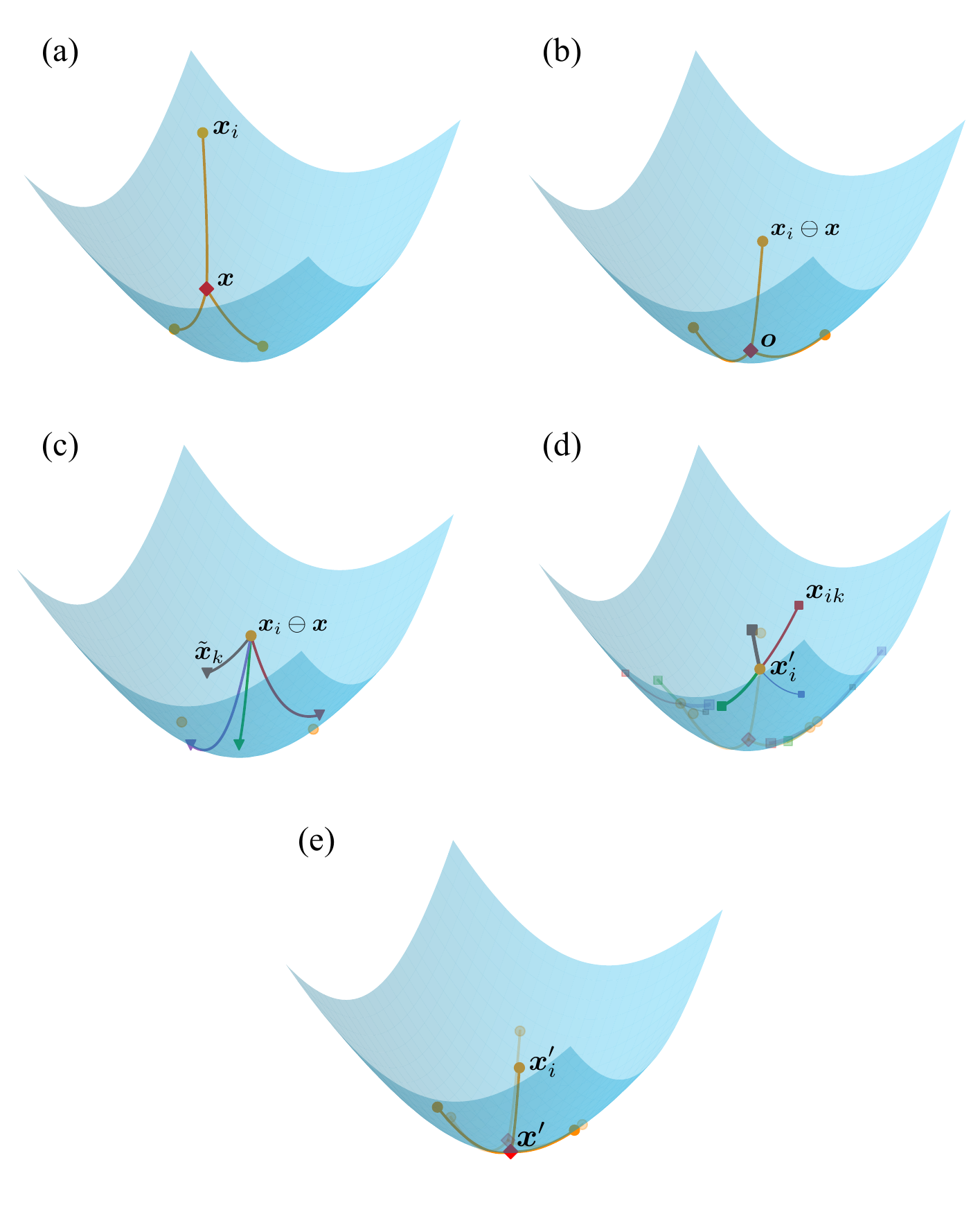}
    \caption{Illustration of HKConv. In this example, there are four kernel points and an input feature $\vx$ with three neighbors in $\gN(\vx)$. (a) The original positions of $\vx$ (the diamond) and $\gN(\vx)$ (the circle points). (b) $\{\vx\} \cup \gN(\vx)$ have been moved to the origin by the parallel transport. (c) The kernel points $\tilde{\vx}_k$ (the down-triangles) and the geodesics between $\vx_i \ominus \vx$ and the kernel points; the correlations  are the lengths of the geodesics. (d) The ${\rm HLinear}$ transformed $\vx_{ik}$'s (the squares), whose weighted centroid is $\vx'_i$; the thickness of the geodesic represents the weight. (e) $\vx' = {\rm HKConv}(\vx)$, the centroid of $\vx'_i$'s.}
    \label{fig:agg}
\end{figure}

Analogous to convolution in the Euclidean space, HKConv applies the kernel points to local neighborhoods of input hyperbolic features. Such neighborhoods are not necessarily formed according to hyperbolic distances. For instance, when the hyperbolic features are embedding of graph features, we can consider the 1-hop neighbors according to graph topology.
Let $\sX \subset \sL^m$ denote the collection of input hyperbolic features. For each $\vx \in \sX$, we use $\gN(\vx) \subset \sX - \{\vx\}$ to denote the neighbors of $\vx$. 
We describe the HKConv layer in the following three steps of operations: first, transformation of relative input features; second, aggregation based on correlation with kernels; third, final aggregation with optional attention. Figure \ref{fig:agg} presents an illustration and we next describe the details.

\paragraph{Step 1.} To ensure the locality of the extracted information, we need to use relative features between an input $\vx$ and its neighbors. More specifically, we need to perform a hyperbolic ``translation''. To this end, we first transform each $\vx_i \in \gN(\vx)$ into the tangent space by applying the logarithmic map, then move them towards $\vo$ by applying the parallel transport ${\rm PT}_{\vx \to \vo}$, and lastly bring them back to $\sL^m$ by applying the exponential map. For convenience, we extend the parallel transport ${\rm PT}_{\vx \to \vy}: \gT_{\vx} \sL^m \to \gT_{\vy} \sL^m$ to ${\rm T}_{\vx \to \vy}: \sL^m \to \sL^m$, which can be regarded as a translation operator defined on the hyperbolic space. Specifically, given $\vx, \vy, \vu \in \sL^m$,
\begin{equation}\label{eq:def_pt_space}
    {\rm T}_{\vx \to \vy}(\vu) := \exp_\vy \left( {\rm PT}_{\vx \to \vy} ( \log_\vx (\vu) ) \right).
\end{equation}
Also, to analogize to ``translation by $\vx$'', we denote
\begin{equation}\label{eq:def_translation}
    \vu \ominus \vx := {\rm T}_{\vx \to \vo} (\vu).
\end{equation}
Clearly, $\vx \ominus \vx = \vo$. Moreover, because logarithmic and exponential maps are distance preserving, \eqref{eq:def_translation} preserves the relative distance between $\vx$ and its neighbors. That is,
\begin{equation*}
    d_\gL(\vx, \vx_i) = d_\gL(\vo, \vx_i \ominus \vx),~ \text{for all}~ \vx_i \in \gN(\vx).
\end{equation*}
We remark that \eqref{eq:def_translation} is similar to ``adding bias'' in some designs of hyperbolic linear layers \citep{ganea2018hyperbolic, chami2019hyperbolic}.

For each $\vx_i \in \gN(\vx)$, the translation $\vx_i \ominus \vx$ gives the relative input feature of $\vx_i$ with respect to $\vx$. The first step of HKConv is to extract information using a learnable hyperbolic linear layer \citep{chen2021fully} for each kernel point $\tilde{\vx}_k$, $k=1,\cdots,K$. 
We denote it by $\operatorname{HLinear}_k: \sL^m \to \sL^n$ and extract 
\begin{equation}
    \vx_{ik} = \operatorname{HLinear}_{k}(\vx_i \ominus \vx), \quad k = 1, \cdots, K. \label{eq:def_hkconv_1}
\end{equation}

\paragraph{Step 2.} Similar to Euclidean convolution, we take the correlation between the input features and the kernel points $\tilde{\vx}_k$. However, unlike the Euclidean domain where the kernels are moved to the neighborhood of the input, when we perform convolution, we move $\gN(\vx)$ to the neighborhood of $\vo$. This, on one hand, avoids different extents of distortion of the kernel points at different locations, and on the other hand ensures simple forms of the expressions of both the exponential and logarithmic maps since the spatial component of the origin is zero. Specifically, each neighbor $\vx_i \in \gN(\vx)$ is moved via the translation $\vx_i \ominus \vx$, and the correlation with the kernel point $\tilde{\vx}_k$ is given by the hyperbolic distance  $d_{\mathcal{L}}(\vx_{i}\ominus \vx, \tilde{\vx}_k)$. 
Unlike Euclidean convolution, in our design, this correlation is used as the weights for aggregating the $\vx_{ik}$ obtained in \eqref{eq:def_hkconv_1}. 
In the hyperbolic space, one way of aggregating features is to find their hyperbolic centroid \citep{law2019lorentzian}, whose output feature is in the same hyperbolic space. More specifically, given a set of input features $\{\vu_i\}_{i=1}^N$ and their corresponding weights $\{\nu_i\}_{i=1}^N$, the hyperbolic centroid (denoted by $\operatorname{HCent}$) of $\{\vu_i\}_{i=1}^N$ is given by
\begin{equation*}
 \operatorname{HCent}(\{\vu_i\}_{i=1}^N,\boldsymbol{\nu}) = 
 \frac{\sum_{i=1}^N\nu_i\boldsymbol{u}_i}{\sqrt{-\kappa}\left|\|\sum_{i=1}^N\nu_i\boldsymbol{u}_i\|_{\mathcal{L}}\right|}.
\end{equation*}

Therefore, aggregating the features yields an output feature for each neighbor of $\vx$:
\begin{equation}
    \vx_{i}' = \operatorname{HCent}(\{\vx_{ik}\}_{k=1}^K, \{d_{\mathcal{L}}(\vx_{i}\ominus \vx, \tilde{\vx}_k)\}_{k=1}^K). \label{eq:def_hkconv_2}
\end{equation}

In \eqref{eq:def_hkconv_2}, the correlation $d_{\mathcal{L}}(\vx_{i}\ominus \vx, \tilde{\vx}_k)$ endows $\vx_{ik}$ with information of local geometry in the feature space.
On the contrary, other hyperbolic message passing \citep{liu2019hyperbolic, chami2019hyperbolic} or attention \citep{gulcehre2018hyperbolic} methods only use graph structural information, but do not take into account the geometry of the feature embedding space.

\paragraph{Step 3.} Different input features may have different numbers of neighbors. The last step in HKconv is to aggregate all the $\vx'_i$'s from the neighborhood. When there is no other information, we simply apply equal weights to all the neighbors. Nevertheless, it is also possible to apply the attention regime so that the weights are learnable. That is,
\begin{equation}
    \vx' = \operatorname{HCent}(\{\vx_i'\}_{i=1}^N, \{w_{i}\}_{i=1}^N), \label{eq:def_hkconv_3} 
\end{equation}
where $\{w_{i}\}_{i=1}^N$ is either taken to be $\boldsymbol{1}_N$ or learned.

Altogether, for each input feature $\vx \in \sL^m$, our hyperbolic convolution produces a new hyperbolic feature $\vx' \in \sL^n$ by cascading the operations \eqref{eq:def_hkconv_1}--\eqref{eq:def_hkconv_3}. 
The learnable parameters are contained in ${\rm HLinear}_k$'s in \eqref{eq:def_hkconv_1}.
For convenience, we call the chain of operations \eqref{eq:def_hkconv_1}--\eqref{eq:def_hkconv_3} an HKConv layer, and denote $\vx' = {\rm HKConv}(\vx; \gN(\vx))$.
When $\gN(\vx)$ is clear from the context, we also directly denote $\vx' = {\rm HKConv}(\vx)$. 

\subsection{Properties of HKConv}

In the above manner, \eqref{eq:def_hkconv_1} and \eqref{eq:def_hkconv_2} analogize to the Euclidean convolution, whereas \eqref{eq:def_hkconv_3} constitutes a local pooling operation. 
Since the aggregation only depends on the neighborhood, which is unaffected by any matrix representation, HKConv is equivariant to permutation of the order of data points. This property is not guaranteed if the convolution is done by applying hyperbolic linear layers to concatenated features, as done in HNN++, where the order of concatenation matters. We summarize this property in the following proposition.

\begin{proposition}[Equivariance to permutation of the order of data points]
Let $\mX \in \R^{N \times (m+1)}$ be the data matrix whose $j$-th row $\vx_j \in \R^{m+1}$ is identified as a point in $\sL^m$. For any matrix $\mP \in \R^{N \times N}$ that permutes the rows of $\mX$,
\begin{equation*}
    {\rm HKConv}(\mP \mX) = \mP {\rm HKConv}(\mX),
\end{equation*}
where HKConv is applied to each row of $\mX$.
\end{proposition}

The above result indicates that HKConv does not depend on a specific Euclidean representation. As to the hyperbolic features themselves, HKConv also enjoys invariance under local translations in the hyperbolic space. 
Specifically, given an HKConv layer, if we perform a parallel translation from an input point $\vx$ along the geodesic in the hyperbolic space, the output $\operatorname{HKConv}(\vx)$ will not change. This result holds locally, i.e., for a given $\vx$ and its neighbors. We formulate it as the following theorem. %

\begin{theorem}[Local translation invariance]\label{thm:trans_invariance}
Fix $\vx \in \sX$ and its neighborhood $\gN(\vx) \subset \sX$. For any $\vy \in \sL^n$ in the geodesic from $\vo$ to $\vx$,
\begin{align}
    &{\rm HKConv}({\rm T}_{\vx \to \vy} (\vx); {\rm T}_{\vx \to \vy}(\gN(\vx))) \nonumber \\
    =~& {\rm HKConv}(\vx; \gN(\vx)). \label{eq:trans_inv}
\end{align}
\end{theorem}

\begin{proof}
For any $\vx_i \in \gN(\vx)$, according to the definition in (4),
\begin{align*}
    &{\rm T}_{\vx \to \vy}(\vx_i) \ominus {\rm T}_{\vx \to \vy}(\vx)\\
    =~& \exp_\vy \left( {\rm PT}_{\vx \to \vy} ( \log_\vx (\vx_i) ) \right) \ominus \vy \\
    =~& {\rm T}_{\vy \to \vo} \left( \exp_\vy \left( {\rm PT}_{\vx \to \vy} ( \log_\vx (\vx_i) ) \right) \right) \\
    =~& \exp_\vo \left( {\rm PT}_{\vy \to \vo} \left( \log_\vy \left( \exp_\vy \left( {\rm PT}_{\vx \to \vy} ( \log_\vx (\vx_i) ) \right) \right) \right) \right) \\
    =~& \exp_\vo \left( {\rm PT}_{\vy \to \vo} \left( {\rm PT}_{\vx \to \vy} ( \log_\vx (\vx_i) ) \right) \right) \\
    =~& \exp_\vo \left( {\rm PT}_{\vx \to \vo} ( \log_\vx (\vx_i) ) \right) \\
    =~& \vx_i \ominus \vx.
\end{align*}

Consequently, the input of ${\rm HLinear}_k$, $k=1,\cdots,K$, in (6) is invariant under the operator ${\rm T}_{\vx \to \vy}$. This implies that $\vx'$ in (8) is also invariant under ${\rm T}_{\vx \to \vy}$, since it only depends on $\{\vx_i \ominus \vx: \vx_i \in \gN(\vx) \}$ and the fixed kernel $\{\tilde{\vx}_k\}_{k=1}^K$. We have thus proved (9).
\end{proof}

Theorem \ref{thm:trans_invariance} indicates that HKConv is indeed an operation that depends on local geometry. Along the geodesics, as long as the relative positions between $\vx$ and its neighbors stay constant, the output of HKConv will not change, regardless of the distance between $\vx$ and the hyperbolic origin.

\subsection{Hyperbolic Kernel Networks}

From the above discussion, HKConv can be defined for any collection of input hyperbolic features with a specified neighborhood for each feature. One possible approach is to choose the nearest neighbors according to the hyperbolic distance. If the data are represented in a graph, then the graph edges can be inherited to the hyperbolic space and used to naturally define neighbors. In particular, unlike HNN++ \citep{shimizu2021hyperbolic}, our hyperbolic convolution allows heterogeneous structures, where different features can have different numbers of neighbors. 

A hyperbolic neural network can be constructed by cascading a number of HKConv layers, which we call a {H}yperbolic {K}ernel {N}etwork (HKN) for simplicity. Depending on the data type, the first layer may be an embedding layer, where data points in the original domain are represented in the hyperbolic space. The output of the final HKConv layer may also need to go through further layers depending on the specific task. For instance, in graph classification where node features are embedded in the hyperbolic space, we apply a global pooling by taking a global centroid so that one single hyperbolic feature is obtained for each graph. For classification tasks, to represent the likelihood of the classes, the output needs to be either a scalar or a Euclidean vector. To this end, we apply the hyperbolic distance layer \citep{liu2019hyperbolic} that maps from $\mathbb{L}^n$ to $\mathbb{R}^\ell$. Specifically, given an input $\vx \in \sL^n$, the output is the vector of distances between $\vx$ and $\ell$ learnable points $\{\boldsymbol{c}_i\}_{i=1}^\ell \subset \mathbb{L}^n$, that is, $[d_{\mathcal{L}}(\boldsymbol{x}, \boldsymbol{c}_1), \cdots, d_{\mathcal{L}}(\boldsymbol{x}, \boldsymbol{c}_\ell)]^\T \in \R^\ell$.

\begin{table*}[t]
    \centering
    \caption{Graph classification results for the chemical and social graphs. Mean accuracy (\%) and standard deviation are reported. We maintain the same level of precision with one decimal place when taking the results of WLHN from \citep{nikolentzos2023weisfeiler}. }

\begin{tabular}{ccccccc} 
\toprule
                               &         & PTC                 & ENZYMES             & PROTEIN             & IMDB-B              & IMDB-M               \\ 
\midrule
\multirow{5}{*}{Data Statistics}                       &     \# of graphs    & 344                 & 600                 & 1113                & 1000                & 1500                 \\
                    &     \# of classes     & 2                   & 6                   & 2                   & 2                   & 3                    \\
                  &    Avg \# of nodes      & 25.56               & 32.63               & 39.06               & 19.77               & 13                   \\
                 &    Avg \# of edges       & 25.96               & 62.14               & 72.82               & 96.53               & 65.94                \\
                &      Hyperbolicity ($\delta$)    & 0.725               & 1.15                & 1.095               & 0.2385              & 0.1157               \\ 
\midrule
\multirow{3}{*}{Euclidean GNN} & GCN \citep{kipf2017semi}    & 63.87$\pm$2.65          & 66.39$\pm$6.91          & 74.54$\pm$0.45          & 73.32$\pm$0.39          & 50.27$\pm$0.38           \\
                               & GIN \citep{xu2018how}     & 66.58$\pm$6.78          & 59.79$\pm$4.31          & 70.67$\pm$1.08          & 72.78$\pm$0.86          & 47.91$\pm$1.03           \\
                               & GMT \citep{baek2021accurate}    & 65.89$\pm$2.16          & 67.52$\pm$4.28          & 75.09$\pm$0.59          & 73.48$\pm$0.76          & 50.66$\pm$0.82           \\ 
\midrule
\multirow{4}{*}{Hyperbolic}    & HGCN \citep{chami2019hyperbolic}    & 55.17$\pm$3.21          & 53.63$\pm$5.12          & 68.41$\pm$2.15          & 61.71$\pm$0.97          & 50.12$\pm$0.71           \\
                               & H2H-GCN \citep{dai2021hyperbolic} & 66.14$\pm$3.91          & 60.84$\pm$4.72          & 72.12$\pm$3.63          & 68.19$\pm$0.82          & 48.31$\pm$0.63           \\ 
                               & HyboNet \citep{chen2021fully} & 65.56$\pm$4.19          & 56.82$\pm$5.39          & 65.36$\pm$2.81          & 71.26$\pm$1.28          & 54.35$\pm$0.98           \\
                               & WLHN \citep{nikolentzos2023weisfeiler} & -          & 62.5$\pm$5.0          & 75.9$\pm$1.9         & 73.4$\pm$3.7         & 49.7$\pm$3.6          \\
\midrule
\multirow{1}{*}{Ours}          & HKN     & \textbf{73.69}$\pm$4.81 & \textbf{82.46}$\pm$5.62 & \textbf{83.22}$\pm$5.19 & \textbf{79.92}$\pm$1.31 & \textbf{56.53}$\pm$1.02  \\
\midrule
\midrule
\multirow{3}{*}{Ablations}     & HKN-direct   & 66.42$\pm$3.13          & 58.31$\pm$4.81          & 64.78$\pm$3.51          & 67.65$\pm$0.89          & 46.31$\pm$0.61           \\
                               & HKN-random   & 61.24$\pm$8.32          & 67.41$\pm$8.62          & 68.85$\pm$7.81          & 68.97$\pm$4.38          & 52.14$\pm$3.83           \\
                               & HKN-train   & NaN                 & NaN                 & NaN                 & NaN                 & NaN                  \\
\bottomrule
\end{tabular}
    \label{tab:gctable}
\end{table*}

\begin{table}[t]
    \centering
    \caption{Graph classification results for the ogbg-molhiv and ogbg-molpcba datasets. Mean (\%) and standard deviation are reported under the standard metrics used in \citep{hu2020ogb}.}
        \begin{tabular}{lcc}
    \toprule
              & \textbf{ogbg-molhiv}     & \textbf{ogbg-molpcba}     \\ 
              & ROC-AUC & Avg. Precision \\
    \midrule          
    GCN \citep{kipf2017semi} & 76.06$\pm$0.97 & 20.20$\pm$0.24 \\ 
    GIN \citep{xu2018how} & 75.58$\pm$1.40 & 22.66$\pm$0.28 \\ 
    HGCN \citep{chami2019hyperbolic} & 75.91$\pm$1.48 & 17.52$\pm$0.20 \\
    HyboNet \citep{chen2021fully} & {76.83}$\pm$0.52 & {20.17}$\pm$0.31 \\
    {WLHN} \citep{nikolentzos2023weisfeiler} & {78.41}$\pm$0.31 & {22.90}$\pm$0.25 \\ 
    \midrule
    HKN & \textbf{79.75}$\pm$0.47 & \textbf{27.87}$\pm$0.32 \\ 
    \bottomrule
    \end{tabular}

    \label{tab:gcltable}
\end{table}

\section{Experiments}

We evaluate the performance of HKN on graph classification in \S\ref{subsec:exp_gc} and node classification in \S\ref{subsec:exp_nc}. In both tasks, each graph node has a hyperbolic representation. 
We present an additional experiment on a non-graph dataset in Appendix \ref{app:mt_dtp}. All the experiments were executed on a GPU server with NVIDIA GeForce RTX 3090 GPUs (24G memory). Each experiment uses a single GPU. The runtime is reported in Appendices \ref{app:time}.

\subsection{Graph Classification}\label{subsec:exp_gc}

In graph classification, one assigns class labels to full graphs. When a graph has a small Gromov hyperbolicity $\delta$, it can be well embedded in the hyperbolic space. In particular, a tree has $\delta = 0$; also, with $\kappa = -1$, $\delta = \tanh^{-1}(1/\sqrt{2}) \approx 0.88$ for $\sL^n$ \citep{sonthalia2020tree}. We expect hyperbolic models to work well for datasets where the graphs have overall similar $\delta$ values. 

\paragraph{Settings} We use the following relatively graph datasets to evaluate our model: chemical graphs including PTC \citep{helma2001predictive}, ENZYMES \citep{schomburg2004brenda}, PROTEIN \citep{borgwardt2005protein}; social graphs including IMDB-B and IMDB-M \citep{yanardag2015deep}. 
In particular, these datasets all have small hyperbolicity (the average $\delta$ over all graphs in each dataset is reported in Table \ref{tab:gctable}). We also test on the large-scale graph classification dataset ogbg-molhiv and ogbg-molpcba \citep{hu2020ogb}. The task is supervised and each dataset is partitioned into a training set, a validation set and a test set.

To build the HKN, we use the validation sets to determine the number of kernel points $K$ from $\{2,3,4,5,6,7,8,9\}$.
To avoid overfitting, dropout is applied to the input of each layer. In all experiments, we use the Riemannian Adam \citep{becigneul2018riemannian} as the optimizer. We consider the standard hyperboloid with curvature $\kappa = -1$ as our underlying hyperbolic space. 
We validate the learning rate from $\{0.001,0.005,0.01\}$; the weight decay rate from $\{0, 0.01, 0.001\}$; and the dropout rate from $\{0.25, 0.5, 0.75, 0.9\}$.

For this task, the HKN architecture contains an embedding layer where graph node features are represented in the hyperbolic space; a cascading of HKConv layers where the  number of layers is validated from $\{2,3,4,5,6,7\}$; a global pooling by taking the centroid of all graph node features; a hyperbolic distance layer whose output represents the likelihood of the classes.

\paragraph{Baselines} For chemical and social graph datesets, we compare HKN with graph neural networks including GCN \citep{kipf2017semi}, GIN \citep{xu2018how}, GMT \citep{baek2021accurate}, and recent hyperbolic neural networks including HGCN \citep{chami2019hyperbolic}, H2H-GCN \citep{dai2021hyperbolic}, HyboNet \citep{chen2021fully}, and WLHN \citep{nikolentzos2023weisfeiler}. For large-scale graph datasets, we compare HKN with the above methods for which relevant results have been reported. We report the mean and the standard deviation from three independent implementations. The results are reported in Tables \ref{tab:gctable} and \ref{tab:gcltable}, respectively. 

\paragraph{Results} HKN consistently achieves the best performance in all the datasets, including both the relatively small datasets and large-scale datasets. Moreover, its advantage over the second best is highly significant in most datasets (approximately 8\% in PTC, 15\% in ENZYMES, 8\% in PROTEIN, 6\% in IBDB-B; 5\% in ogbg-molpcba). We also note that the hyperbolic neural network baselines do not show a clear advantage over graph neural networks. This indicates that these baselines do not extract effective geometric information other than the graph structure. 
We believe the considerable improvement reflects that HKN has the capacity in extracting useful patterns from local geometric embedding of features in full graphs.

\begin{table}[t]
    \centering
    \caption{Graph classification results with different numbers of kernel points. Mean accuracy (\%) and standard deviation are reported.}
    \resizebox{\columnwidth}{!}{
    \begin{tabular}{cccccc} 
\toprule
  & PTC                 & ENZYMES             & PROTEIN             & IMDB-B              & IMDB-M               \\ 
\midrule
2 & 66.73$\pm$4.21          & 56.05$\pm$5.22          & 63.60$\pm$4.00          & 66.83$\pm$1.29          & 53.36$\pm$0.95           \\
3 & 60.12$\pm$4.24          & 59.78$\pm$5.60          & 66.83$\pm$3.82          & 70.02$\pm$1.53          & 54.48$\pm$0.97           \\
4 & 60.50$\pm$4.63          & \textbf{82.46}$\pm$5.62 & \textbf{83.22}$\pm$5.19 & \textbf{79.92}$\pm$1.31 & 55.51$\pm$0.99           \\
5 & 60.58$\pm$4.22          & 80.92$\pm$5.21          & 80.12$\pm$3.82          & 76.67$\pm$1.40          & 53.77$\pm$0.84           \\
6 & \textbf{73.69}$\pm$4.81 & 76.25$\pm$5.27          & 74.15$\pm$4.10          & 69.37$\pm$1.40          & 55.13$\pm$1.53           \\
7 & 62.74$\pm$4.34          & 80.12$\pm$5.65          & 68.61$\pm$3.89          & 77.46$\pm$1.55          & \textbf{56.53}$\pm$1.02  \\
8 & 62.86$\pm$4.30          & 80.42$\pm$5.81          & 66.50$\pm$4.55          & 65.15$\pm$1.13          & 51.87$\pm$1.17           \\
9 & 67.28$\pm$4.35          & 82.27$\pm$5.31          & 66.44$\pm$3.89          & 79.13$\pm$1.22          & 54.33$\pm$0.93           \\
\bottomrule
\end{tabular}
    }
    \label{tab:kptable}
\end{table}

\begin{figure}[t]
    \centering
    \includegraphics[width=.8\linewidth]{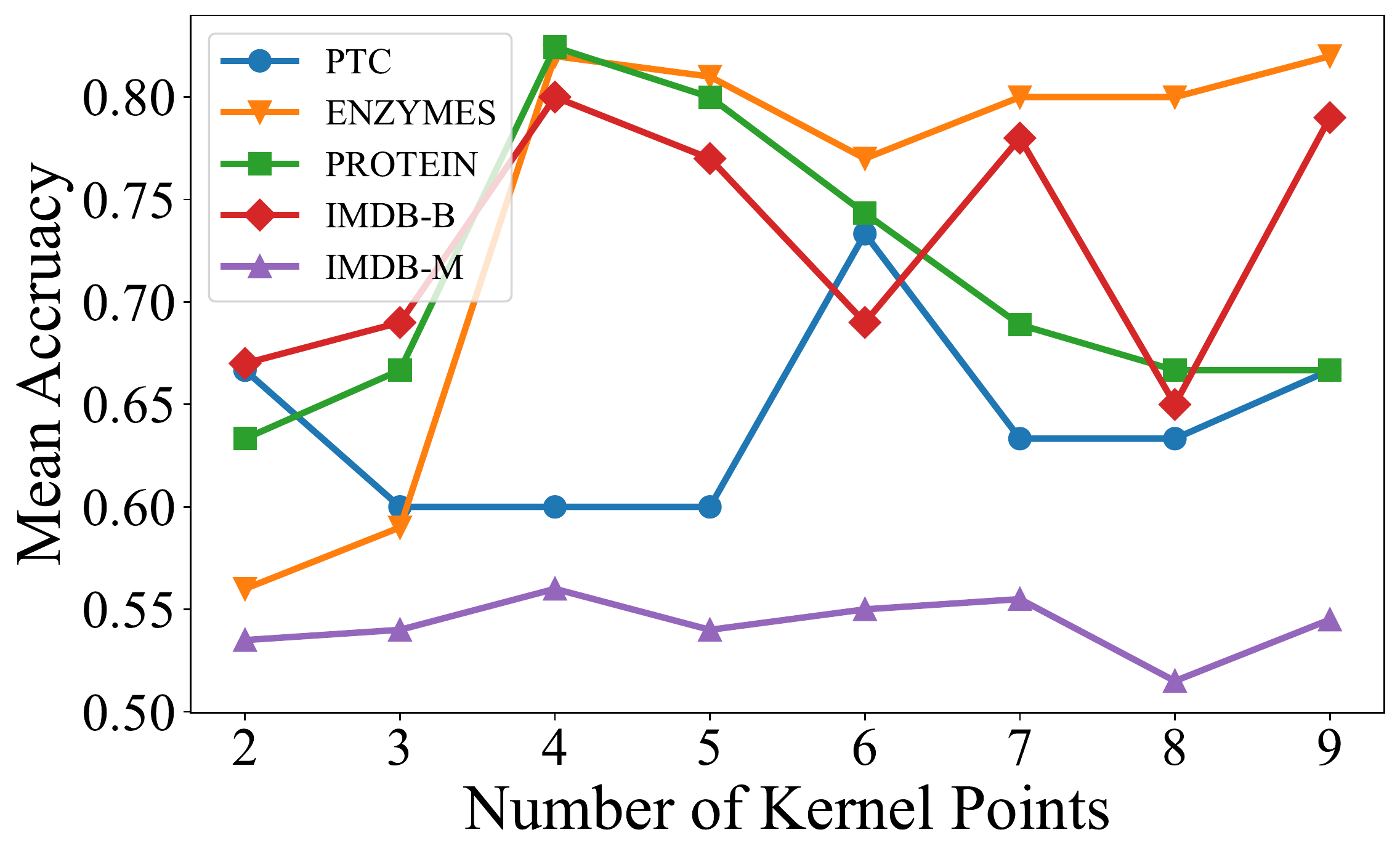}
    \caption{Graph classification with different numbers of kernel points.}
    \label{fig:gckp}
\end{figure}

\begin{table*}[t]
\centering
\caption{Node classification results. Mean F1 score (\%) and standard deviation are reported. NaN indicates a numerically unstable result.}

\begin{tabular}{ccccccccc} 
\toprule
               & Cornell             & Texas               & Wisconsin           & Chameleon           & Squirrel            & Actor               & Cora                & Pubmed               \\ 
\midrule
\# of Nodes    & 183                 & 183                 & 251                 & 2,277               & 5,201               & 7,600               & 2,708               & 19,717               \\
\# of Edges    & 280                 & 295                 & 466                 & 31,421              & 198,493             & 26,752              & 5,278               & 44,327               \\
\# of Features & 1,703               & 1,703               & 1,703               & 2,325               & 2,089               & 931                 & 1,433               & 500                  \\
\# of Classes  & 5                   & 5                   & 5                   & 5                   & 5                   & 5                   & 6                   & 3                    \\
Hyperbolicity  & $\delta=1$          & $\delta=1$          & $\delta=1$          & $\delta=1.5$        & $\delta=1.5$        & $\delta=1.5$        & $\delta=11$         & $\delta=3.5$        \\ 
\midrule
MLP            & 81.89$\pm$6.40          & 80.81$\pm$4.75          & 85.29$\pm$3.31          & 46.21$\pm$2.99          & 28.77$\pm$1.56          & 36.53$\pm$0.70          & 75.69$\pm$2.00          & 87.16$\pm$0.37           \\
\midrule
GCN \citep{kipf2017semi}           & 60.54$\pm$5.30          & 55.14$\pm$5.16          & 51.76$\pm$3.06          & 64.82$\pm$2.24          & 53.43$\pm$2.01          & 27.32$\pm$1.10          & 86.98$\pm$1.27          & 88.42$\pm$0.50           \\
GAT  \citep{velickovic2018graph}           & 61.89$\pm$5.05          & 52.16$\pm$6.63          & 49.41$\pm$4.09          & 60.26$\pm$2.50          & 40.72$\pm$1.55          & 27.44$\pm$0.89          & 87.30$\pm$1.10          & 86.33$\pm$0.48           \\
GraphSAGE  \citep{hamilton2017inductive}   & 75.95$\pm$5.01          & 82.43$\pm$6.14          & 81.18$\pm$5.56          & 58.73$\pm$1.68          & 41.61$\pm$0.74          & 34.23$\pm$0.99          & 86.90$\pm$1.04          & 88.45$\pm$0.50           \\
GCNII   \citep{chen2020simple}       & 77.86$\pm$3.79          & 77.57$\pm$3.83          & 80.39$\pm$3.40          & 63.86$\pm$3.04          & 38.47$\pm$1.58          & 37.44$\pm$1.30          & \textbf{88.37}$\pm$1.25 & \textbf{90.15}$\pm$0.43  \\
\midrule
Geom-GCN   \citep{pei2019geom}    & 60.54$\pm$3.67          & 66.76$\pm$2.72          & 64.51$\pm$3.66          & 60.00$\pm$2.81          & 38.15$\pm$0.92          & 31.59$\pm$1.15          & 85.35$\pm$1.57          & 89.95$\pm$0.47           \\
WRGAT  \citep{suresh2021breaking}         & 81.62$\pm$3.90          & 83.62$\pm$5.50          & 86.98$\pm$3.78          & 65.24$\pm$0.87          & 48.85$\pm$0.78          & 36.53$\pm$0.77          & 88.20$\pm$2.26          & 88.52$\pm$0.92           \\
LINKX   \citep{lim2021large}        & 77.84$\pm$5.81          & 74.60$\pm$8.37          & 75.49$\pm$5.72          & 68.42$\pm$1.38          & 61.81$\pm$1.80          & 36.10$\pm$1.55          & 84.64$\pm$1.13          & 87.86$\pm$0.77           \\
GloGNN    \citep{li2022finding}     & 83.51$\pm$4.26          & 84.32$\pm$4.15          & 87.06$\pm$3.53          & 69.78$\pm$2.42          & 57.54$\pm$1.39          & 37.35$\pm$1.30          & 88.31$\pm$1.13          & 89.62$\pm$0.35           \\
GloGNN++ \citep{li2022finding}      & \textbf{85.95}$\pm$5.10 & 84.05$\pm$4.90          & 88.04$\pm$3.22          & 71.21$\pm$1.84          & 57.88$\pm$1.76          & 37.70$\pm$1.40          & 88.33$\pm$1.09          & 89.24$\pm$0.39           \\ 
\midrule
HGCN   \citep{chami2019hyperbolic}        & 79.43$\pm$0.47          & 70.13$\pm$0.32          & 83.26$\pm$0.51          & NaN                 & 62.31$\pm$0.57          & 36.58$\pm$0.79          & 79.84$\pm$0.27          & 80.41$\pm$0.42           \\
H2H-GCN  \citep{dai2021hyperbolic}        & 75.52$\pm$0.82          & 71.47$\pm$0.63          & 88.71$\pm$0.82          & 78.71$\pm$0.96          & 66.85$\pm$0.72          & 42.73$\pm$0.86          & 83.74$\pm$0.87          & 82.26$\pm$0.83           \\
HyboNet  \citep{chen2021fully}      & 77.27$\pm$0.71          & 72.23$\pm$0.94          & 86.52$\pm$0.51          & 74.91$\pm$0.58          & 69.07$\pm$0.64          & 45.74$\pm$0.82          & 81.42$\pm$0.93          & 78.45$\pm$1.17           \\
WLHN  \citep{nikolentzos2023weisfeiler}      & 77.29 $\pm$ 4.66          & 75.41 $\pm$ 5.98          & 78.62 $\pm$ 3.44 & -          & 55.76 $\pm$ 0.92          & 36.42 $\pm$ 1.42          & -          & -                     \\
\midrule
HKN    (Ours)      & 84.14$\pm$0.53          & \textbf{90.94}$\pm$0.66 & \textbf{91.31}$\pm$0.36 & \textbf{85.27}$\pm$0.57 & \textbf{75.26}$\pm$0.38 & \textbf{67.20}$\pm$0.68 & 80.34$\pm$0.59          & 77.62$\pm$0.54           \\
\bottomrule
\end{tabular}
    \label{tab:nc}
\end{table*}

\paragraph{Ablations} We further analyze the effect of the important components in HKConv, namely, the way we perform translation in \eqref{eq:def_hkconv_1} and the fixed hyperbolic kernel points that minimize \eqref{eq:loss_ker}.  We consider the following three ablations:
\begin{itemize}[leftmargin=12pt, parsep=0pt]
    \item \textbf{HKN-direct}: we apply hyperbolic linear layers directly to each $\vx_i$ in \eqref{eq:def_hkconv_1} instead of $\vx_i \ominus \vx$, and use $d_\gL(\vx_i, {\rm PT}_{\vo \to \vx}(\tilde{\vx}_k))$ to translate the kernels to the neighborhood of $\vx$ instead of $d_{\mathcal{L}}(\vx_{i}\ominus \vx, \tilde{\vx}_k)$ in \eqref{eq:def_hkconv_2}.
    \item \textbf{HKN-random}: we randomly sample the kernel points $\{\tilde{\vx}_k\}_{k=1}^K$, independently from a wrapped normal distribution \citep{nagano2019wrapped} with unit standard deviation, instead of solving \eqref{eq:loss_ker}.
    \item \textbf{HKN-train}: we regard the positions of the kernel points as trainable parameters, which are optimized together with other network parameters.
\end{itemize}
We apply these alternatives to the chemial and social graph classification tasks and present the results in the same Table \ref{tab:gctable}. Indeed, the performance deteriorates significantly if we perform translation by moving the kernels to the neighborhood of input features. We believe that HKN-direct fails to capture some important geometric information due to different distortion of the kernel at different locations. Similarly, we get low accuracy if we randomly generate the kernel points. This result validates the formulation of the loss function in \eqref{eq:loss_ker}. Moreover, if we consider the locations of the kernel points as learnable parameters, the training process will be numerically unstable and produce NaN for all datasets. The ablation studies have validated the necessity and competitiveness of the key components of HKConv.

\paragraph{Analysis on number of kernels} In the above experiment, we determined the number of kernel points by validation. Intuitively, there is tradeoff between expressivity and overfitting.
In the following, we study how the performance of HKN varies with the different numbers of kernels to further validate the contribution of the kernel points. We record the performance using HKN with $K=\{2,3,4,5,6,7,8,9\}$ kernel points. To focus on the number of kernels, the other hyperparameters are fixed according to the main task. The results are reported in Table \ref{tab:kptable} and plotted in Figure \ref{fig:gckp}. 
In Figure \ref{fig:gckp}, we only include the mean but not the standard deviation for clarity.

The results reveal that indeed, a moderate number of kernel points leads to the best performance. We observe that when the number of kernels is $K=2$, the performance is on par with the baseline methods. This implies that using only few kernels cannot fully extract the hyperbolic pattern. On the other hand, if we keep increasing the number of kernel points, the performance will not keep improving. With a large number of kernel points, the neural network becomes complex and may overfit the data.

\subsection{Node Classification}\label{subsec:exp_nc}

\begin{table*}[t]
\centering
\caption{Node classification results with different numbers of kernel points. The mean F1 score (\%) and standard deviation are reported.}
    \begin{tabular}{ccccccccc} 
\toprule
  & Cornell             & Texas               & Wisconsin           & Chameleon           & Squirrel            & Actor               & Cora                & Pubmed               \\ 
\midrule
2 & 82.08$\pm$0.44          & 79.08$\pm$0.36          & 87.20$\pm$0.75          & 83.08$\pm$0.58          & 75.11$\pm$0.42          & 65.87$\pm$0.55          & \textbf{80.34}$\pm$0.59 & \textbf{77.62}$\pm$0.54  \\
3 & 76.71$\pm$0.47          & \textbf{90.94}$\pm$0.66 & 88.90$\pm$0.33          & 79.79$\pm$0.50          & 75.03$\pm$0.50          & 65.14$\pm$0.45          & 78.97$\pm$0.48          & 77.27$\pm$0.27           \\
4 & {83.92}$\pm$0.31 & 84.19$\pm$0.90          & \textbf{91.31}$\pm$0.36 & 80.84$\pm$0.42          & \textbf{75.26}$\pm$0.38 & 66.96$\pm$0.34          & 80.16$\pm$0.44          & 77.06$\pm$0.56           \\
5 & 76.87$\pm$0.34          & 88.50$\pm$0.24          & 88.82$\pm$0.44          & 83.20$\pm$0.72          & 74.14$\pm$0.25          & 66.52$\pm$0.64          & 78.55$\pm$0.74          & 76.33$\pm$0.29           \\
6 & 72.03$\pm$0.62          & 81.77$\pm$0.27          & 83.19$\pm$0.48          & \textbf{85.27}$\pm$0.57 & 74.62$\pm$0.38          & 65.90$\pm$0.83          & 78.27$\pm$0.74          & 75.27$\pm$0.49           \\
7 & 74.75$\pm$0.46          & 86.64$\pm$0.24          & 81.87$\pm$1.29          & 82.51$\pm$0.40          & 74.39$\pm$0.54          & 63.96$\pm$0.43          & 77.90$\pm$0.65          & 75.94$\pm$0.36           \\
8 & \textbf{84.14}$\pm$0.53          & 84.06$\pm$0.34          & 88.73$\pm$0.63          & 82.55$\pm$0.43          & 74.82$\pm$0.37          & \textbf{67.20}$\pm$0.68 & 78.10$\pm$1.16          & 75.03$\pm$0.29           \\
9 & 75.11$\pm$0.39          & 86.43$\pm$0.46          & 89.65$\pm$0.57          & 78.95$\pm$0.64          & 74.62$\pm$0.61          & 66.10$\pm$0.56          & 76.27$\pm$0.97          & 70.05$\pm$0.62           \\
\bottomrule
\end{tabular}
    \label{tab:nckp}
\end{table*}

In node classification, one classifies nodes within a graph according to some nodes with known labels. Unlike graph classification, no global pooling is taken in node classification. The final output is a vector of distances for each node, representing the likelihood of all the classes.

\paragraph{Settings} As discussed in \S \ref{subsec:exp_gc}, we expect HKN to work well for graph datasets whose Gromov hyperbolicity $\delta$ \citep{sonthalia2020tree} is similar to that of a hyperboloid. We test our HKN on the following datasets: the {WebKB} datasets \citep{craven2000learning} including Cornell, Texas and Wisconsin, where nodes are web pages and edges are hyperlinks; the {Wikipedia Network} datasets \citep{rozemberczki2021multi} including Chameleon and Squirrel, where nodes are Wikipedia pages and edges are links between two pages; the {Actor Co-occurrence Network} dataset \citep{tang2009social} (Actor), where nodes represent actors and two nodes are connected if they co-appear in the same Wikipedia page. In these datasets, the hyperbolicity $\delta$ is either $1$ or $1.5$, very close to that of a hyperboloid. Moreover, we also test HKN on well-known citation networks including Cora \citep{namata2012query} and PubMed \citep{sen2008collective}, which show larger (and thus weaker) hyperbolicities.

\paragraph{Baselines} We compare HKN with the following baseline methods: (1) multi-layer perceptron (MLP); (2) general graph neural networks including GCN \citep{kipf2017semi}, GAT \citep{velickovic2018graph}, GraphSAGE \citep{hamilton2017inductive}, GCNII \citep{chen2020simple};
(3) methods for heterophilous graphs including Geom-GCN \citep{pei2019geom}, WRGAT \citep{suresh2021breaking}, LINKX \citep{lim2021large}, GloGNN and GloGNN++ \citep{li2022finding}; (4) hyperbolic GCN methods: HGCN \citep{chami2019hyperbolic}, H2H-GCN \citep{dai2021hyperbolic}, HyboNet \citep{chen2021fully}. We report in Table \ref{tab:nc} the mean F1 score (\%) and standard deviation from three independent runs.

\paragraph{Results} Our HKN achieves statistically significant improvement over other methods on five out of six benchmarks (Texas, Wisconsin, Chameleon, Squirrel, Actor) which show strong hyperbolicity. On Cornell, its performance is also very close to the best result. In particular, HKN consistently excels the other hyperbolic methods by a large margin. On the other hand, general graph neural networks are better than hyperbolic ones in Cora and PubMed, which only show very weak hyperbolicity. In these non-hyperbolic datasets, our method is still comparable to other hyperbolic neural networks.

\begin{figure}[t]
    \centering
    \includegraphics[width = \columnwidth]{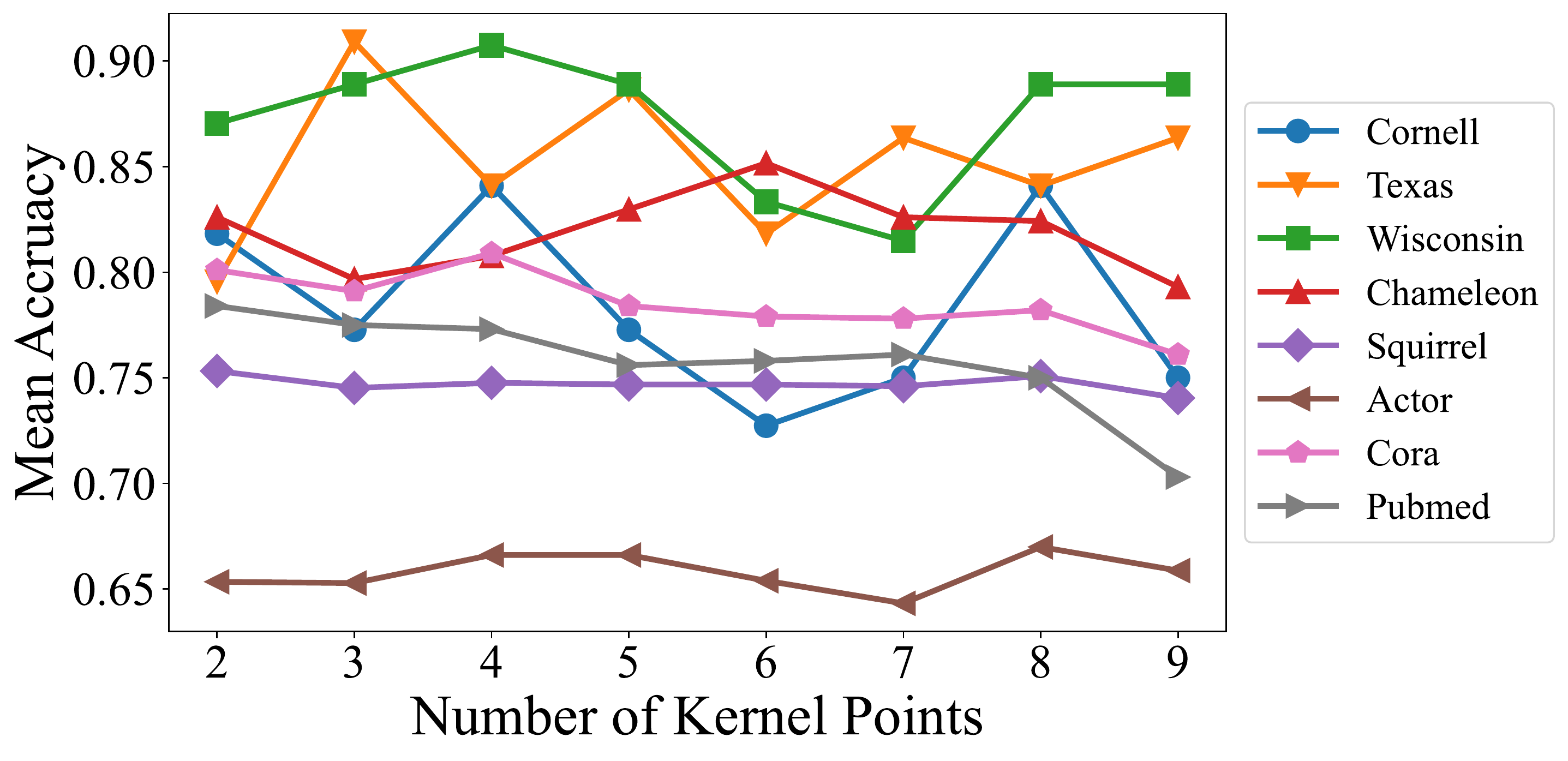}
    \vskip -0.1in
    \caption{Node classification results with different numbers of kernel points.}
    \label{fig:nckp}
    \vskip -0.15in
\end{figure}

\paragraph{Analysis on number of kernels} We also study how the performance of HKN changes with different numbers of kernels. Fixing the other hyperparameters, we record the performance of HKN using $K = \{2,3,\cdots,9\}$ kernel points. The results are displayed in Table \ref{tab:nckp} and plotted in Figure \ref{fig:nckp}.

We notice that for datasets with smaller hyperbolicity (the first six columns), the best performance is achieved with a moderate $K$, which accommodates the tradeoff between expressivity and overfitting. However, for Cora and PubMed, which are less hyperbolic, increasing the number of kernels will not enhance the performance and the best performance is already achieved with $K=2$ kernels. We conclude that the kernel points in HKConv are effective in extracting hyperbolic features, while non-hyperbolic data may not benefit from more kernel points.

\subsection{Machine Translation and Dependency Tree Probing}\label{app:mt_dtp}

\begin{table*}[ht]
    \centering
    \caption{Machine translation and dependency tree probing results. The BLEU (BiLingual Evaluation Understudy) scores on the test set are reported for machine translation. UUAS, Dspr., Root\%, Nspr. are reported for dependency tree probing.}
    {\small

\begin{tabular}{ccccccccc} 
\toprule
            & \multicolumn{4}{c}{Machine Translation}                       & \multicolumn{4}{c}{Dependency Tree Probing}                  \\ 
\cmidrule{2-9}
            & IWSLT' 14     & \multicolumn{3}{c}{WMT' 14}                   & \multicolumn{2}{c}{Distance}  & \multicolumn{2}{c}{Depth}    \\ 
\cmidrule{2-9}
       & d=64          & d=64          & d=128         & d=256         & UUAS          & Dspr.         & Root\%      & Nspr.          \\ 
\midrule
ConvSeq2Seq \citep{gehring2017convolutional} & 23.6          & 14.9          & 20.0          & 21.8          & -             & -             & -           & -              \\
Transformer \citep{ott2018scaling} & 23.0          & 17.0          & 21.7          & 25.1          & 0.36          & 0.3           & 12          & 0.88           \\ 
DynamicConv \citep{chen2020dynamic} & 24.1      & 15.2 & 20.4  & 22.1 & -             & -             & -           & -         \\
BiBERT \citep{qin2022bibert}     & 23.8      & 18.7 & 22.5  & 27.8 & -             & -             & -           & -         \\
\midrule
HNN++ \citep{shimizu2021hyperbolic}      & 22.0          & 17.0          & 19.4          & 21.8          & -             & -             & -           & -              \\
HATT \citep{gulcehre2018hyperbolic}        & 23.7          & 18.8          & 22.5          & 25.5          & 0.5           & 0.64          & 49          & 0.88           \\
HyboNet \citep{chen2021fully}    & 25.9          & 19.7          & 23.3          & 26.2          & 0.59          & 0.7           & 64          & 0.92           \\
HKN (Ours)        & \textbf{27.3} & \textbf{20.1} & \textbf{25.6} & \textbf{29.1} & \textbf{0.62} & \textbf{0.74} & \textbf{72} & \textbf{0.94}  \\
\bottomrule
\end{tabular}
}
    \label{tab:mt}
\end{table*}

In addition to the tasks for graph data introduced before, we consider a non-graph task. In this task, we consider language data that show hierarchical structures. Machine translation translates one language to another and is a sequence-to-sequence modeling task. Dependency tree probing then analyzes the obtained model. 

\paragraph{Settings} For fair comparison, we use the network architecture of the hyperbolic transformer model used by \citet{chen2021fully}, except that we apply HKConv for the aggregation in the following manner. Given the query set $\mathcal{Q}=\{\vq_1,\ldots,\vq_{|\mathcal{Q}|}\}$, the key set $\mathcal{K}=\{\vk_1,\ldots,\vk_{|\mathcal{K}|}\}$, and the value set $\mathcal{V}=\{\vv_1,\ldots,\vv_{|\mathcal{V}|}\}$, where $|\mathcal{K}|=|\mathcal{V}|$, We implement HKConv for each $i = 1, \cdots, \vert \gV \vert$ as follows: 
\begin{align*}
\vv_{ijk}&=\operatorname{HLinear}_{k}(\vv_j \ominus \vv_i), \quad k = 1, \cdots, K; \\
\vv_{ij}'&=\operatorname{HCent}(\{\vv_{ijk}\}_{k=1}^K, \{d_{\mathcal{L}}(\vv_{j}\ominus \vv_i, \tilde{\vx}_k)\}_{k=1}^K); \\
\bm{\mu}_i&=\operatorname{HCent}(\{\vv'_{ij}\}_{j=1}^{|\mathcal{V}|}, \{w_{ij}\}_{j=1}^{|\mathcal{V}|})),
\end{align*}
where for $j = 1,\cdots,\abs{\gV}$, $$w_{ij}=\frac{\exp \left(\frac{-d_{\mathcal{L}}^{2}\left(\bm{q}_{i}, \bm{k}_{j}\right)}{\sqrt{n}}\right)}{\sum_{k=1}^{|\mathcal{K}|} \exp \left(\frac{-d_{\mathcal{L}}^{2}\left(\bm{q}_{i}, \bm{k}_{k}\right)}{\sqrt{n}}\right)}.$$
Here, the first two equations are exactly \eqref{eq:def_hkconv_1} and \eqref{eq:def_hkconv_2}, presented for clear notation. The last equation is \eqref{eq:def_hkconv_3} where the weights are determined by the query set and the key set. 
Overall, HKConv performs a further transformation of $\vv_i$ that utilizes the local geometric embedding.

Following \citet{chen2021fully}, we take two machine translation datasets: IWSLT'14 and WMT'14\footnote{We correct the typo of the year number in the reference.} English-German. Dependency tree probing is tested on the former.

\paragraph{Baselines} We compare with ConvSeq2Seq \citep{gehring2017convolutional}, Transformer \citep{ott2018scaling}, DynamicConv \citep{chen2020dynamic}, BiBERT \citep{qin2022bibert}; as well as hyperbolic methods including HNN++ \citep{shimizu2021hyperbolic}, HATT \citep{gulcehre2018hyperbolic} and HyboNet \citep{chen2021fully}. Note particularly that HNN++ uses their hyperbolic convolution which, while not applicable in the graph classification tasks due to the heterogeneous graph structures, naturally works in this task. 
We report the results in Table \ref{tab:mt}, where we implement our method and DynamicConv and BiBERT. The benchmark results for ConvSeq2Seq, Transformer, HNN++, HATT and HyboNet are taken from \citet{shimizu2021hyperbolic} and \citet{chen2021fully}.

\paragraph{Results} HKN consistently excels all the baseline methods in both tasks and in all the metrics. Since we have intentionally kept the same attention regime as HyboNet but replacing aggregation with HKConv, our better performance does not result from attention. We believe the encouraging results are due to the better expressivity of HKConv as well as the capability of learning local information to countervail the global attention in the transformer.

\section{Related Works}

\paragraph{Hyperbolic Neural Networks} The earliest hyperbolic network was HNN \citep{ganea2018hyperbolic}, where linear layers and recurrent layers were defined according to the Poincar\'{e} ball model. HNN was recently generalized to HNN++ \citep{shimizu2021hyperbolic}, which introduced concatenation, convolution and attention mechanisms in the Poincar\'{e} ball. \citet{nickel2018learning} pointed out the numerical instability of operations in the Poincar\'{e} ball and promoted the use of the Lorentz model. More recently, \citet{chen2021fully} proposed to perform operations directly in the hyperbolic space without relying on the tangent spaces and thus derived the fully hyperbolic network. \citet{fan2022nested} took a similar approach, but constrained the maps to be Lorentz transformations. It is also possible to use hybrid models. For instance, \citet{gulcehre2018hyperbolic} established a hyperbolic attention using both the Lorentz and the Klein models. For a more complete list of various hyperbolic networks with their underlying models, we refer to the recent survey by \citet{peng2021hyperbolic}.

\paragraph{Non-Euclidean Convolution} Recently, CNNs have been generalized to different manifolds. Geometric convolutional models include the spherical CNN \citep{cohen2018spherical}, the mesh CNN \citep{hanocka2019meshcnn} and the icosahedral CNN \citep{cohen2019gauge}. In these works, the input features are defined as a function on the space, not embedded in the space, and thus their methods do not generalize to hyperbolic models as considered in our work. 

Convolution has also been generalized to graph data, with a focus on effective processing of graph topology. Specifically, GCN convolution is implemented by local message passing and aggregation \citep{kipf2017semi}. Due to the tree-like structure and power law distribution in many graph data, it is natural to use the hyperbolic space to represent geometric information missing from graph topology \citep{yang2022hyperbolic}.
Many hyperbolic networks \citep{liu2019hyperbolic, chami2019hyperbolic, bachmann2020constant, dai2021hyperbolic, zhang2021lorentzian} generalized GCN convolution. In these works, the input features are transformed directly by various hyperbolic fully-connected layers and the aggregation inherits from the graph edge weights. Unlike them, our HKConv transforms relative features in each neighborhood and therefore encodes more local geometric information when applied to graph data.

We also remark that the title of a recent work \citep{lensink2022fully} seems to suggest similar network construction. However, there is no direct relation between our works since in that work, ``hyperbolic'' describes the type of differential equation and ``convolution'' refers to the regular operation for images. 

\section{Conclusion and Limitation}

We have introduced HKConv, a novel hyperbolic convolutional layer based on point kernels that extract information from local geometric relations between input features. 
The corresponding hyperbolic network, HKN, has achieved state-of-the-art performance in both graph-level and node-level tasks for tree-like datasets. We have particularly showed the advantage of the important components in our convolution.

One possible limitation of this work is that the number of parameters in HKConv grows with the number of kernels, which may lead to overfitting for small datasets when the number of kernels is not small. To address this issue, a future working direction is to explore sampling regimes when building the kernels for reducing model complexity.

\bibliographystyle{IEEEtranN}
\bibliography{bib_file}

\ifCLASSOPTIONcaptionsoff
  \newpage
\fi

\begin{IEEEbiography}[{\includegraphics[width=1in,height=1.25in,clip,keepaspectratio]{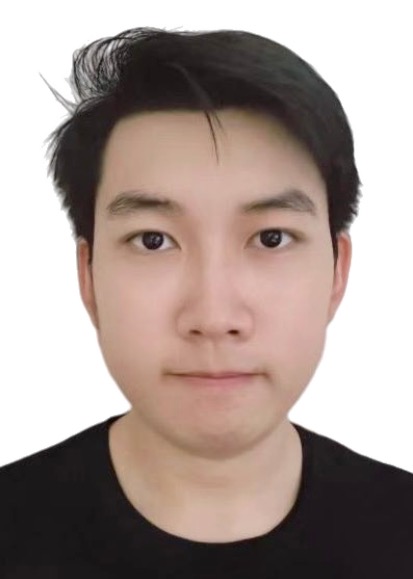}}]{Eric Qu}
received the B.S.~degree in data science from the Duke Kunshan University in 2023. He is currently working toward the Ph.D.~degree in Computer Science at University of California, Berkeley. His research interests include geometric machine learning and AI for Science.
\end{IEEEbiography}

\vspace{11pt}

\begin{IEEEbiography}[{\includegraphics[width=1in,height=1.25in,clip,keepaspectratio]{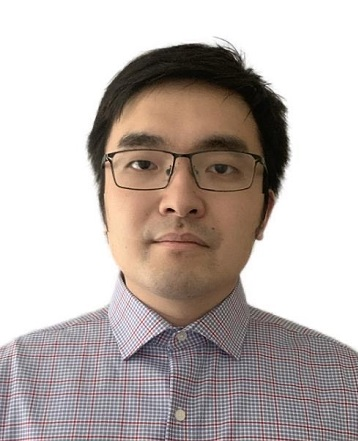}}]{Dongmian Zou}  (Member,~IEEE) received the B.S.~degree in Mathematics (First Honour) from
the Chinese University of Hong Kong in 2012 and the Ph.D.~degree in Applied Mathematics and Scientific Computation from the University of Maryland, College Park in 2017. 

From 2017 to 2020, he served as a post-doctorate researcher at the Institute for Mathematics and its Applications, and the School of Mathematics at the University of Minnesota, Twin Cities. He joined Duke Kunshan University in 2020 where he is currently an Assistant Professor of Data Science in the Division of Natural and Applied Sciences. He is also affiliated with the the Zu Chongzhi Center for Mathematics and Computational Sciences (CMCS) and the Data Science Research Center (DSRC). His research is in the intersection of applied harmonic analysis, machine learning and signal processing.
\end{IEEEbiography}

{\appendices 

\begin{figure*}[hbt!]
    \centering
    \includegraphics[width=0.48\linewidth]{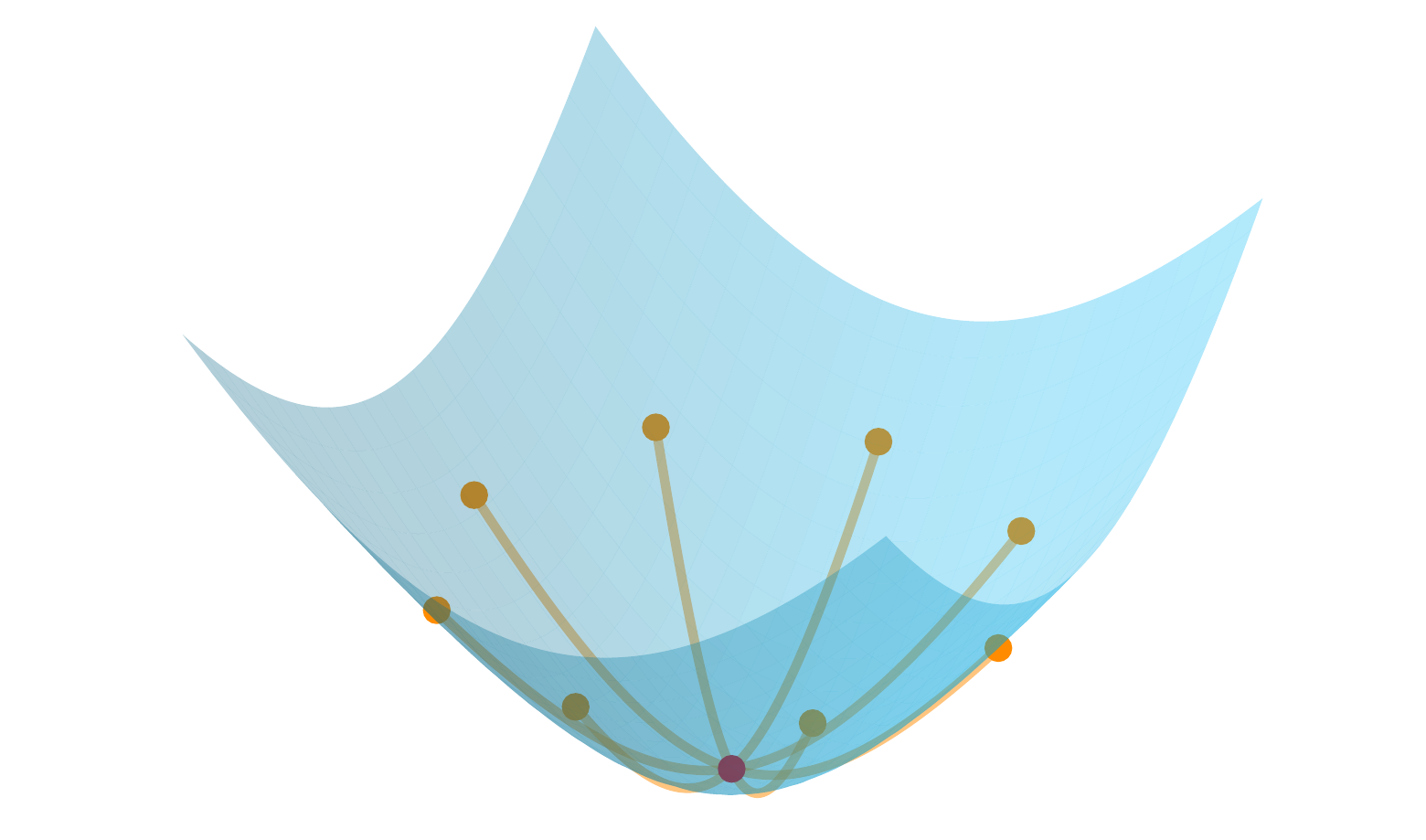}
    \includegraphics[width=0.48\linewidth]{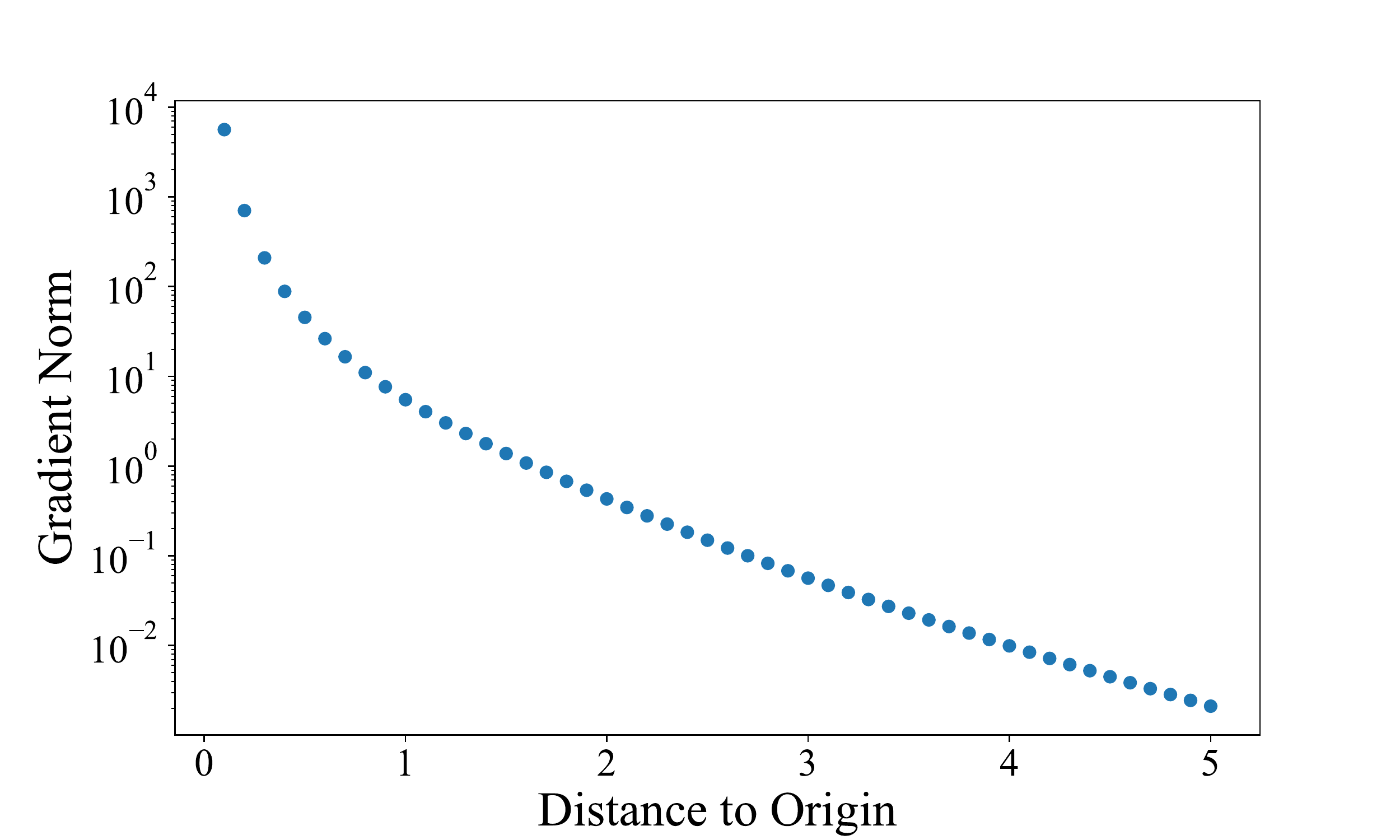}
    \caption{Left: illustration of the sampled points used in the experiment. Right: relationship between distances to origin and gradient norms.}
    \label{fig:kpex1}
\end{figure*}

\begin{table*}[hbt!]
\centering
\caption{Runtime of different hyperbolic GCNs in graph-related tasks.}
\label{tab:time}
\centering
\begin{tabular}{cccccc} 
\toprule
Task                                  & Dataset   & HKNet  & HyboNet         & H2H-GCN         & HGCN   \\ 
\midrule
\multirow{5}{*}{Graph classification (\S \ref{subsec:exp_gc})} & PTC       & 1.04s  & \textbf{0.55}s   & 0.63s            & 0.97s   \\
                                      & ENZYMES   & 2.48s  & 1.66s            & \textbf{1.43}s   & 2.27s   \\
                                      & PROTEINS  & 3.05s  & 2.26s            & \textbf{2.17}s   & 2.94s   \\
                                      & IMDB-B    & 2.82s  & 1.83s            & \textbf{1.64}s   & 2.62s   \\
                                      & IMDB-M    & 5.63s  & 2.92s            & \textbf{2.58}s   & 4.21s   \\ 
\midrule
\multirow{8}{*}{Node classification (\S \ref{subsec:exp_nc})} & Cornell   & 0.029s & \textbf{0.0066}s & 0.0089s          & 0.027s  \\
                                      & Texas     & 0.021s & \textbf{0.0076}s & 0.0084s          & 0.025s  \\
                                      & Wisconsin & 0.020s & \textbf{0.0076}s & 0.0091s          & 0.031s  \\
                                      & Chameleon & 0.035s & 0.016s           & \textbf{0.013}s  & 0.031s  \\
                                      & Squirrel  & 0.15s  & 0.016s           & \textbf{0.014}s  & 0.043s  \\
                                      & Actor     & 0.051s & \textbf{0.012}s  & 0.014s           & 0.042s  \\
                                      & Cora      & 0.024s & \textbf{0.0084}s & 0.011s           & 0.035s  \\
                                      & Pubmed    & 0.022s & 0.0082s          & \textbf{0.0076}s & 0.027s  \\
\bottomrule
\end{tabular}

\end{table*}

\section{Additional Results}

\subsection{More on Hyperbolic Kernel Points}\label{app:hkp}

In this section, we show the necessity of including the second term in \eqref{eq:loss_ker}, i.e., not allowing the kernel points to be far from the origin. As mentioned in the main text, gradients can vanish if the kernel points have large norms. We adopt the following experiment show this phenomenon. We uniformly sample $K=8$ points along the unit circle in $\sL^2$, as illustrated in Figure \ref{fig:kpex1}. Then, we translate them along the geodesics from the origin, in the manner that all the points have the same distance to the origin. For each distance value, we record the gradient norm of the first term of \ref{eq:loss_ker}, i.e. the reciprocals of the pairwise distances. The results are shown in Figure \ref{fig:kpex1}. Clearly, the gradient norm decrease exponentially with the distance, which shows a typical case where gradients vanish when the distance between the kernel points and the origin is large.

\subsection{Runtime}\label{app:time}

We show runtime for training one epoch of typical hyperbolic networks in Table \ref{tab:time}. Our method is slower than HyboNet and H2H-GCN because of its complexity which depends on $K$. Still, the complexity of our method is comparable to HGCN. The speed is compensated by consistently and significantly better performance in all the tasks.

\clearpage

}

\end{document}